\documentclass{article}
\usepackage{graphicx} 

\usepackage[dvipsnames]{xcolor}
\usepackage{caption}
\usepackage{tabularx}
\usepackage{fvextra}
\usepackage{placeins}
\usepackage[
backend=biber,
style=numeric,
]{biblatex}
\usepackage{booktabs}
\usepackage{array}
\usepackage{tikz}
\usetikzlibrary{shapes.geometric, arrows, positioning, calc}

\title{Scrambled text: training Language Models to correct OCR errors using synthetic data}
\author{Jonathan Bourne}
\date{September 2024}

\begin{document}

\maketitle

\begin{abstract}
OCR errors are common in digitised historical archives significantly affecting their usability and value.
Generative Language Models (LMs) have shown potential for correcting these errors using the context provided by the corrupted text and the broader socio-cultural context, a process called Context Leveraging OCR Correction (CLOCR-C). However, getting sufficient training data for fine-tuning such models can prove challenging. This paper shows that fine-tuning a language model on synthetic data using an LM and using a character level Markov corruption process can significantly improve the ability to correct OCR errors. Models trained on synthetic data reduce the character error rate by 55\% and word error rate by 32\% over the base LM and outperform models trained on real data. Key findings include; training on under-corrupted data is better than over-corrupted data; non-uniform character level corruption is better than uniform corruption; More tokens-per-observation outperforms more observations for a fixed token budget. The outputs for this paper are a set of 8 heuristics for training effective CLOCR-C models, a dataset of 11,000 synthetic 19th century newspaper articles and \verb|scrambledtext| a python library for creating synthetic corrupted data.
\end{abstract}

\section{Introduction}
\label{sect:intro}
Optical Character Recognition (OCR) is a valuable tool in the historical archive digitisation process. OCR allows the text of archived material to be digitised at character level, allowing archival documents to be searched and manipulated like more modern natively digital documents. However, performing OCR on archival documents is not without errors \cite{smith_research_2018}. These errors can be particularly acute in archival media such as newspapers and periodicals due to their complex layouts \cite{chiron_impact_2017}. Such errors impact the value of the digitisation process as the ability to search for individuals or specific places and events can degrade dramatically \cite{traub_impact_2015, chiron_impact_2017,  hill_quantifying_2019}.

As a result of the issues with the OCR processes, post-OCR correction has become an area of active research \cite{chiron_icdar2017_2017,  nguyen_survey_2021, neudecker_survey_2021}. More recently, with advances in machine learning and the development of the transformer architecture \cite{vaswani_attention_2017}, the use of transformer-based language models for post-OCR correction has grown in popularity \cite{thomas_leveraging_nodate, boros_post-correction_2024, bourne_clocr-c_2024}. Transformers offer additional benefits over more traditional approaches to OCR correction \cite{nguyen_survey_2021, neudecker_survey_2021}, in that they can be prompted to have a base context for the text (`This is a receipt', `This is a poem'), as well as being able to use the Transformers facility to handle long-range textual dependencies to provide context on corrupted characters and words. Such an approach to OCR correction can be called Context Leveraging OCR Correction (CLOCR-C). Pre-trained language models have been shown to reduce errors by over 60\% \cite{bourne_clocr-c_2024}. However, the cost of using them can be prohibitively high for large projects. An alternative approach is to fine-tune smaller open-source models \cite{thomas_leveraging_nodate, dereza_have_2024} specifically on the task of CLOCR-C.

Fine-tuning is a process of performing additional training on a pre-trained LM to make it better at a specific task \cite{zhao_lora_2024}. However, the process can be expensive and time-consuming. Recent research suggests that much of the functionality of transformer-based LMs exists on a relatively small sub-space of the LM parameters and that by fine-tuning a small subset of parameters on a specific task, the model can experience significant improvements in that task using much less compute. These processes are collectively known as Parameter Efficient Fine Tuning (PEFT)\cite{fu_effectiveness_2023, han_parameter-efficient_2024}. One of the most popular approaches to PEFT is Low-Rank Adaptors (LoRA) \cite{zhao_lora_2024}, which, when combined with other recent advances, such as floating point quantization \cite{dettmers_llmint8_2022}, substantially reduces the cost of fine-tuning, due to reduced GPU use, whilst also providing regularization to prevent overfitting \cite{biderman_lora_2024}. For example, Quantized LoRa Llama2 7B required RAM drops from 38.7 to 7.5, and token throughput increases 20\% relative to the base Llama2 model \cite{zheng_llamafactory_2024}.

The Fine-tuning process is dependent on having data in sufficient quantity and quality to be able to improve the performance of the model, whilst there are several appropriate datasets available \cite{evershed_correcting_2014,  chiron_icdar2017_2017, rigaud_icdar_2019, jiang_prototype_2022, booth_bln600_2024, bourne_clocr-c_2024}, this does not mean they have the necessary qualities to act as training material. Recent research has highlighted that noisy or poor-quality data can impact the performance of trained LMs \cite{bagla_noisy_2024, li_can_2023, li_textbugger_2019,huang_noisyag-news_2024, naplava_understanding_2021}, such noise is a problem when it comes to creating ground truth datasets for OCR correction (as noted by \cite{evershed_correcting_2014}). One way to avoid this problem is using expert transcribed datasets \cite{evershed_correcting_2014, booth_bln600_2024, bourne_clocr-c_2024}. However, transcription is time-consuming and or costly, and using previously transcribed data may not have the correct error distribution or sufficient corruption \textcite{rychalska_models_2019} to be of value.

An alternative to transcribed data is to use synthetic data created using a noising or corruption process \cite{edunov_understanding_2018, rychalska_models_2019, naplava_understanding_2021}, as previous work has shown that OCR errors have identifiable patterns \cite{nguyen_deep_2019}.
Recent advances in training LMs suggests that synthetic data can play a crucial role in training and fine-tuning LM models due to the guaranteed high-quality nature of the data \cite{gunasekar_textbooks_2023, li_textbooks_2023, tan_15-pints_2024}. However, synthetic data needs to be applied carefully, as there is evidence that when LMs are trained on recursively generated synthetic data, they can suffer model collapse and produce gibberish \cite{shumailov_ai_2024}. Whilst synthetic data has been shown to be valuable in various LM applications, little is known about how such an approach would affect the final performance of LMs in CLOCR-C.

This paper contributes to the literature by demonstrating that synthetically generated text, corrupted using a learned function can improve the performance of LMs to correct OCR errors. Specifically, it answers the following questions.
\begin{itemize}
    \item Can synthetic corrupted OCR data be used to train a Langauge model such that it is better able to perform CLOCR-C?
    \item How does the level of corruption of the synthetic data impact the performance of the fine-tuned language model?
    \item What are a set of basic guidelines for practitioners using synthetic data for fine-tuning a CLOCR-C LM?
\end{itemize}

The rest of the paper is as follows, Section \ref{sect:data} introduces the dataset used in this paper, Section \ref{sect:Method} describes the methods for creating the synthetic text and the experiments used to measure its effectiveness as training data, Section \ref{sect:results} reports the results of the experiments, Section \ref{sect:disc} is the discussion interpreting the findings and their place relative to previous literature, Section \ref{sect:recs} provides a list of heuristics for training CLOCR-C models, finally Section \ref{sect:concs} provides the conclusions and considers future work

\section{Data}
\label{sect:data}

When fine-tuning LMs for OCR correction, data selection must be done carefully. This is because much open-source data has already been used to train the most high-performing LMs, which can lead to the memorisation of the text and apparently exceptional recovery rates due to data leakage. As a result, much potential training data is ruled out, such as \cite{jiang_prototype_2022} as it is based on data from the Gutenberg project \cite{hart_project_1971}, which is used in training many LMs. As such, this paper will use 4 OCR correction datasets transcribed from archive newspapers, which it is believed have not been used as training data on language models. Using archival newspaper data is also interesting due to the high levels of corruption that can tend to exhibit and the fact that the data will all be from a similar distribution regarding style and structure. The datasets used are described below.

\begin{itemize}
    \item NCSE: This is a 40,000-word corpus \cite{bourne_clocr-c_2024} made up of 91 articles from the original Nineteenth-Century Serials Edition.
    \item SMH: These are articles from the Sydney Morning Herald containing 52,000 words and 159 articles created by \cite{evershed_overproof_2014} using the TROVE dataset \cite{holley_many_2009}.
    \item CA: Chronicling America, an 18,000-word dataset across 46 articles by \cite{evershed_overproof_2014}
    \item BLN600: A collection of 600 articles from 19th century English newspapers from \cite{booth_bln600_2024}, this paper will also use the sequence level dataset created by \cite{thomas_leveraging_nodate}.
\end{itemize}

Of these 4 datasets, the NCSE dataset will be used as the test dataset as it has the highest overall error rate and is an excerpt from a much larger dataset that would benefit from CLOCR-C. The other three datasets will be used to train the corruption function.

\FloatBarrier
\section{Method}
\label{sect:Method}
The method section is broken into 5 parts. The first section describes the synthetic data generation process, followed by the evaluation and metrics used in the paper. Once the evaluation metrics are described, the Markov process used to corrupt the data is defined. Then, the Llama LM architecture and training parameters are introduced and explained. Finally, the experiments are described. 

\begin{figure}[htbp]
    \centering
    \textbf{Process Flow Diagram}
    \begin{tikzpicture}[
        node distance = 1cm and 2cm,
        box/.style = {rectangle, draw, text width=2.5cm, text centered, minimum height=1cm, fill=white},
        arrow/.style = {->, thick}
    ]

    \node[rectangle, draw, fill=green!30, text width=2.5cm, text centered, minimum height=1cm] (scrambled) {Scrambled text};
    \node[rectangle, draw, fill=SkyBlue!30, text width=2.5cm, text centered, minimum height=1cm, left=of scrambled] (ocr) {OCR aligned text};
    \node[rectangle, draw, fill=SkyBlue!30, text width=2.5cm, text centered, minimum height=1cm, right=of scrambled] (params) {Synthetic text parameters};
    \node[rectangle, draw, fill=pink!50, text width=2.5cm, text centered, minimum height=1cm, below=of params] (synthetic) {Generate synthetic texts};
    \node[rectangle, draw, fill=pink!50, text width=2.5cm, text centered, minimum height=1cm, below=of synthetic] (subset) {Subset texts};
    \node[rectangle, draw, fill=pink!50, text width=2.5cm, text centered, minimum height=1cm, below=of ocr] (probdist) {Get prob distribs};
    \node[rectangle, draw, fill=pink!50, text width=2.5cm, text centered, minimum height=1cm, below=of scrambled] (corrupt) {Corrupt text};
    \node[rectangle, draw, fill=SkyBlue!30, text width=2.5cm, text centered, minimum height=1cm, below=of probdist] (target) {Target CER};
    \node[rectangle, draw, fill=green!30, text width=2.5cm, text centered, minimum height=1cm, below=of corrupt] (train) {Fine-tuned model};
    \node[rectangle, draw, fill=green!30, text width=2.5cm, text centered, minimum height=1cm, below=of subset] (base) {Base LLaMA};
    \node[rectangle, draw, fill=pink!50, text width=2.5cm, text centered, minimum height=1cm, below=of train] (evaluate) {Evaluate performance};
    \node[rectangle, draw, fill=SkyBlue!30, text width=2.5cm, text centered, minimum height=1cm, left=of evaluate] (test) {Test data};

    \draw[arrow] (scrambled) -- (probdist);
    \draw[arrow] (scrambled) -- (corrupt);
    \draw[arrow] (ocr) -- (probdist);
    \draw[arrow] (params) -- (synthetic);
    \draw[arrow] (synthetic) -- (subset);
    \draw[arrow] (probdist) -- (corrupt);
    \draw[arrow] (subset) -- (corrupt);
    \draw[arrow] (target) -- (corrupt);
    \draw[arrow] (subset) -- (train);
    \draw[arrow] (corrupt) -- (train);
    \draw[arrow] (base) -- (train);
    \draw[arrow] (train) -- (evaluate);
    \draw[arrow] (test) -- (evaluate);

    \end{tikzpicture}

    \caption{How the elements of the project hang together can be visualised as above. In this diagram the data is shown in blue, the processes in pink and the computational tools such as scrambledtext and Llama as green}
    \label{fig:project_flow}
\end{figure}
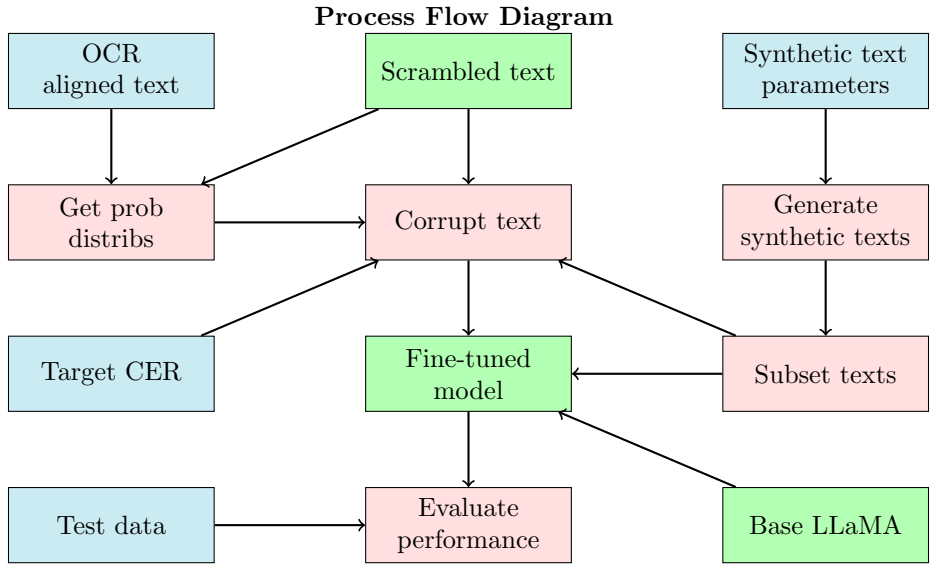

For clarity, Figure \ref{fig:project_flow} shows the flow diagram of the project. The flow diagram has three types: Data, such as the line-aligned texts; Processes, such as corrupting the text; and Computational tools, such as \verb|scrambledtext|. Each element of Figure \ref{fig:project_flow} will be described in detail in the rest of the method.

\FloatBarrier
\subsection{Evaluation}

This paper evaluates the performance of the LMs using the Character Error Rate (CER) and the Word Error Rate, which are calculated using
\begin{equation}
    \textrm{ER} = \frac{S + D + I}{S + D + C}
\end{equation}

Where S is the number of substitutions, D is the number of deletions, I is the number of insertions, and C is the correct total number, the denominator is equivalent to the number of characters in the ground truth reference document. The modifications refer to characters or words depending on whether the metric is CER or WER. 

Having defined the metrics used to measure corruption, we will next introduce the corruption function itself.

\subsection{ScrambledText: Creating the Markov Corruption Process}
The corruption generation function aims to simulate the character corruption type found in OCR documents. Previous research has produced high-quality corruption processes \cite{rychalska_models_2019, naplava_understanding_2021}; however, these focus on typographical or human errors, not OCR-style errors. Other similar approaches, such as back-translation \cite{ott_fairseq_2019}, cannot control the amount of error. Only Genelog \cite{gupte_lights_2021} is explicitly designed to re-create OCR errors, which it does by creating a synthetically corrupted PDF then performing OCR on it. Whilst very thorough, this approach is computationally expensive and technically tricky due to automatically generating appropriate layouts, making it impractical for low-resource projects.

This section is divided into three subsections, the first describes and defines the Markov Model, the second discusses the implementation of the model in \verb|ScrambledText|, and The third discusses the application of the model in this project and provides example text.

\subsubsection{The Markov Model}
\label{sect:markov_model}
\begin{figure}[htbp]
    \centering
    \textbf{The Corruption Network}
        \begin{tikzpicture}[node distance=2cm and 1.5cm,
            every node/.style={draw, shape=ellipse, minimum width=1.8cm, minimum height=1cm, align=center},]
        \node (correct) {Correct};
        \node[below=of correct] (substitute) {Substitute};
        \node[below=of substitute] (finish) {Finish};
        \node[left=of substitute] (insert) {Insert};
        \node[right=of substitute] (delete) {Delete};

        \draw[->,shorten >=1pt] (correct) to [out=0,in=0,looseness=2.5] (finish);
        \draw[->,shorten >=1pt] (correct) to [out=240,in=20] (insert);
        \draw[->,shorten >=1pt] (correct) to [out=270,in=90] (substitute);
        \draw[->,shorten >=1pt] (correct) to [out=300,in=120] (delete);
        \draw[->,shorten >=1pt] (insert) to [out=-60,in=180] (finish);
        \draw[->,shorten >=1pt] (substitute) to [out=180,in=0] (insert);
        \draw[->,shorten >=1pt] (substitute) to [out=270,in=90] (finish);
        \draw[->,shorten >=1pt] (delete) to [out=210,in=30] (finish);

        \draw[->,shorten >=1pt] (insert) to [out=90,in=180,loop,looseness=4.8] (insert);

        \end{tikzpicture}
    \caption{The corruption network is applied at character level to the text; the conditional transition probabilities between the nodes are learnt on a per character basis from parallel OCR and ground truth texts}
    \label{fig:markov-network}
\end{figure}
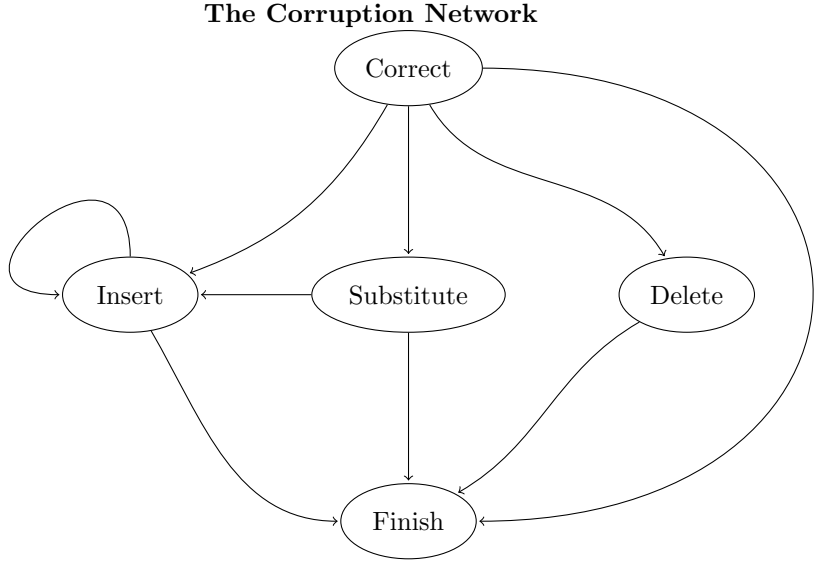

Due to the issues with the existing methods, this paper develops ScrambledText, a simple, fast corruption process based on a Markov model specifically designed to simulate OCR corruption errors using a straightforward training process and can output text with an arbitrary amount of corruption. The Markov Model is shown in Figure \ref{fig:markov-network}; the model takes a single correct character $x$ and outputs a sequence of characters $y$ with a length greater than or equal to 0. Once in the model, each character has a learned conditional probability distribution of passing through the model unaltered, being deleted, substituted, and having a new character inserted after it. The insertion node of the model has a self-referential link, meaning that if a substitution or insertion takes place, the model can keep inserting characters with probability $P(I|x)$. 

The Markov model can be represented mathematically as the conditional probability distribution that $x$ produces output sequence $y$, as shown below.

\begin{equation}
    P(y|x) =P(D|x) + \sum_{i = 1}^\infty \left[ P(S|x) + P(C|x)  \right] P(I|x)^{i-1}
\label{eq:prob_dist}
\end{equation}
Where $P(C|x)$ is the probability of output sequence $y$ being the correct character, $P(D|x)$ is the probability of $x$ being deleted, $P(S|x)$ is the probability of $x$ being substituted for another character, and $P(I|x)$ the probability of a new character being inserted. The self-reference in the network allowing multiple new characters to be inserted means that the input character $x$ can return a sequence $i$ long, where $0\geq i \geq \infty$, the probability of $y$ being 1 or greater is shown using the summation from all possible values of $i$.

Equation \ref{eq:prob_dist} represents the probability distribution mapping only a single character to its output sequence, not the entire document. However, as this approach models all characters as independent, the joint probability of output sequence $Y$ for input sequence $X$ can be represented as

\begin{equation}
P(Y|X) = \prod_{j=1}^{j=n}P(y_j|x_j)
    \label{eq:joint_prob}
\end{equation}

Which is the product of all output sequences for the $n$ input characters in the document.

A significant advantage of the Markov model is the ease with which the overall CER can be modified. As CER$\approx 1-P(C)$, this paper uses a straightforward approach of directly setting the character level conditional probability of being correct based on the target CER, that is $P(C|x) = 1 - \textrm{CER}$. The approach used to renormalise in this paper multiplies the other elements in the row by a factor and then renormalises to ensure the result is a probability distribution. Thus, the adjustment factor is for a conditional probability matrix $M$ with $m$ characters and $n$ states.

\begin{equation}
    f_{ik} = \frac{1-\textrm{P}'(k|i)}{1-\textrm{P}(k|i)}
\end{equation}
Where $\textrm{P}'(k|i)$ is the adjusted conditional probability that the $i$th character has state $k$. This means for each row of the matrix $M$, we can perform the operation

\begin{equation}
    M'_{ij} = \left\{
\begin{array}{ll}
  M'_{ij} \cdot f_{ik} & \textrm{if} \;j \neq k \\
  M_{ij} &\textrm{if} \;j = k
\end{array}
\right.
\end{equation}

This operation scales all elements in row $i$ by the scaling factor created from element $k$. To ensure that the rows are a valid distribution, they are then renormalised using

\begin{equation}
    S_i = \sum^n_{j=1} M_{ij}
\end{equation}

to get the sum of each row and then divide by each element in that row

\begin{equation}
    M''_{ij} = \frac{M'_{ij}}{S_i}
\end{equation}

for all rows in the conditional matrix. Although $k$ can be any state, in this paper, only $P(C|x)$ will be adjusted. This approach can be critiqued as being a somewhat blunt instrument. Still, it is conceptually easy to understand and computationally cheap, making it ideal for this exploration of synthetic data for training LMs to perform CLOCR-C. 

An in-depth exploration of the implications of this corruption model is beyond the scope of this paper. However, it is worth noting that, as the output sequences are a probability distribution,  the observed CER for a uniformly corrupted text will itself have a probability distribution centred approximately on  

\begin{equation}
    P(\textrm{CER}|X) \approx 1 - \sum P(C|x)P(\hat{x}|X)
\end{equation}

which is one minus the joint probability of the characters being correct for input sequence $X$ and observed character distribution $\hat{x}$, which may differ from the character distribution $x$ learned during the training.

\subsubsection{Process Implementation}

The corruption function has been implemented in Python as a library called \verb|scrambledtext|. The process requires only parallel OCR texts that have been line-aligned; the learned conditional corruption probabilities can be easily exported as JSON files and passed between projects. In addition, as the process is based on adjacent symbols, the process is language and script agnostic.
The library has two main classes, \verb|ProbabilityDistributions|, which learns the corruption conditional probability distributions, and \verb|CorruptionEngine|, which uses the learned corruption distributions to corrupt text. The library is lightweight, relying only on Python libraries included in the standard distribution, and is easily extensible using class inheritance.

\subsubsection{Application of the model}

The model is trained using the BLN600, SMH and CA datasets, which are used to try to create diversity in the observed errors. Each dataset's OCR and ground truth texts are character aligned using the \verb|genalog| \cite{gupte_lights_2021}. The texts are then sequentially loaded, and the conditional distributions are learned. Once the distributions have been learned, they are saved as a JSON file for use in the rest of the project. With the conditional probabilities learned, text can now be arbitrarily corrupted. Table \ref{tab:corruption_results} shows an example of text \cite{lovelace_sketch_1842} being corrupted from CER = 0 to CER = 0.5. Table \ref{tab:corruption_results} shows both the Target CER that the corruption function was aiming for and the Observed CER of the output sequence and drift from character insertions. The difference is due to the stochastic nature of the process.

\begin{table*}
\centering
\caption{As can be seen as the corruption function increases the CER, the text becomes increasingly illegible}
\label{tab:corruption_results}
\begin{tabular}{|p{0.1\linewidth}|p{0.14\linewidth}|p{0.7\linewidth}|}
\hline
\textbf{Target CER} & \textbf{Observed CER} & \textbf{Corrupted Text} \\
\hline
0.00 & 0.00 & We may say most aptly that the Analytical Engine weaves algebraical patterns just as the Jacquard-loom weaves flowers and leaves.  \\
\hline
0.10 & 0.09 & We may say mot aptly tAt tho Analytical Engine wedaves Talgebraical patterns just as theJacquard-lom weaves fowrs anld leavesi  \\
\hline
0.20 & 0.20 & We may san mest puly thab th e Analyticel Engrte weFaves ealgebr'airal Fpttens just a ihe J acquard-loofwcves flowers antd lleaves. 1 \\
\hline
0.30 & 0.28 & Wcelay  say mtost apteJy -thnt tha AndasyTlical E1ninewcaves algeibiacalf pattetnsa yus t -as t heo .Jaeqaard-loomweaovesflowers an-d leaves.  \\
\hline
0.40 & 0.47 & ge mnay dsa most arpti tha-ihie A natlyei..cil Ea-gire Twiea1yes  rglbrd;ienl settias ojis as- 'lhe Jtcnad-loom wteaveml lloweia iand Ple-ave   .' \\
\hline
0.50 & 0.62 & W.e may 'sav tosmt aptdy rl1 ltiit  fSe k.nal  riclI-En-ggn  taa-we  ,\textbackslash eragoeibraac1lplate,,lns g"a--  aa.thie oJsciaa.rd.-loo-ml Cvv,evosfl-ower1s  aad ave.  \\
\hline
\end{tabular}
\end{table*}

With the method of corruption defined, the process for generating the synthetic articles, and overall dataset, which will be corrupted is described.

\subsection{Synthetic Data}

The data is created by prompting a Language model (in this case GPT4o) to write a piece of text under a set of guidelines. The prompt is shown in Table \ref{tab:prompt_vars}.

\begin{table}[ht]
\centering
\caption{The prompt is populated by replacing the bold words in curly braces with appropriate descriptors. The descriptors can be found in Table \ref{tab:prompt_vars} }
\begin{tabular}{|p{0.9\textwidth}|} 
\hline
\multicolumn{1}{|c|}{\textbf{Synthetic text generation prompt}} \\ \hline
It is the year \textbf{\{year\}}. Using the text provided below surrounded by triple \#, write a \textbf{\{word\_count\}} word \textbf{\{writing\_style\}} \textbf{\{text\_type\}} with a \textbf{\{sentiment\}} sentiment, the persona of the writer is \textbf{\{persona\}}, the reading level should be \textbf{\{complexity\}}.\\
Note: The resultant text may be distasteful to modern readers, that is ok. Respond only in plain text, do not use markdown\\
\#\#\#\\
\textbf{\{text\}}\\
\#\#\#\\

\\ \hline
\end{tabular}
\end{table}

The seven words inside curly braces, shown in Table \ref{tab:prompt_vars}, are the variables used to create diversity between the texts. The seed prompt, shown as `\{text\}' in the prompt, was obtained by downloading the Wikipedia timeline of the 19th century and the timeline of British diplomacy in the 19th century; combining both lists created xx events to describe. These timelines provide a brief description of an actual 19th-century event, which could then be styled by the rest of the prompt. The variables used for prompt styling were chosen to be appropriate for the 19th century and are shown in table \ref{tab:prompt_vars}; there are 3888 possible combinations of text styling. The timeline and prompt variables were then sampled 11,000 times, and the text generated by GPT-4o. This resulted in a corpus of 11,000 texts with 3.5 million words. To create a more manageable training set, a random substring of text 200 tokens long is removed from each article; this is used to create a 10,000-observation training set where each observation is of consistent length. The remaining 1000 observations are split between a test an validation set, but are not used in this project. An example of a prompt an the resultant article are shown in Supplementary material Section 3, the synthetic data can be downloaded from \cite{bourne_scrambled_2024}.

\begin{table}[h]
\centering
\caption{Variables and options used in the prompt template}
\begin{tabularx}{\textwidth}{lX}
\toprule
\textbf{Variable} & \textbf{Options} \\
\midrule
\textbf{text type} & newspaper article, obituary of a named person, editorial, book excerpt, letter to the editor, personal diary entry \\
\midrule
\textbf{writing style} & formal, informal, satirical, religious, polemic, romantic, persuasive, descriptive \\
\midrule
\textbf{persona} & general public, women's rights, politics, economics and trade, military, reactionary, chartist, clergy, arts and culture \\
\midrule
\textbf{sentiment} & positive, neutral, negative \\
\midrule
\textbf{complexity} & simple, moderate, advanced \\
\bottomrule
\end{tabularx}
\label{tab:prompt_vars}
\end{table}

Having provided an overview of the dataset and evaluation metrics, the models and training parameters used throughout the experiments is introduced.

\subsection{The Llama Model and training parameters}

This paper constrains the number of experiments by not attempting to optimise the model's performance but rather to explore the influence of the training synthetic data on performance. This Data Centric modelling \cite{strickland_andrew_2022, zha_data-centric_2023, salehi_data-centric_2024} approach means that we need only evaluate one model architecture and that the parameters of the model can be fixed so that the impact of changes in the data is clear.

We use the Llama-3.1 8B instruct \cite{dubey_llama_2024} an 8 Billion parameter model which, at the time of writing, was one of the top performing models of its size class. Like most current top-performing models, Llama is an autoregressive (or causal) decoder-only model based on the transformer architecture. In addition, the model has an open licence, making fine-tuning and distribution of the resultant models possible. 

In relation to the discussion of PEFT in Section \ref{sect:intro}, this paper will use LoRA \cite{hu_lora_2021}, specifically Rank Stabilised LoRA \cite{kalajdzievski_rank_2023} which has been shown to produce better results than the original LoRA approach.

The model will be trained using the Huggingface framework and the Unsloth \cite{han_unsloth_2024} implementation of Llama. Training will use the lightning.ai platform, the GPU will be a 24Gb Nvidia L4. Table \ref{tab:hyper_params} shows the most important hyper-parameters used in the training. The hyper-parameters were chosen to ensure that the model would not exceed the 24Gb RAM of the GPU; from this perspective, setting the context length was most important from experiments 1024 was chosen as it would be able to handle the tokens produced by all levels of corruption, detail on this choice can be seen in supplementary material Figure 1. After an appropriate context window was found, the batch size and LoRA parameters could be chosen. For a full list of hyper-parameters, see the code. 

\begin{table}[h]
    \caption{Hyper-parameters used in the model.}
    \centering
    \begin{tabular}{l c}
        \toprule
        \textbf{Hyper-parameter} & \textbf{Value} \\
        \midrule
        LoRA Rank             & 64 \\
        LoRa Alpha            & 32 \\
        Learning rate    & 5e-5 \\
        Epochs           & 1 \\
        Scheduler        & AdamW 8-bit \\
        Batch size       & 16 \\
        Context window   & 1024 \\
        \bottomrule
    \end{tabular}
    \label{tab:hyper_params}
\end{table}

with the experimental framework defined, the experiments themselves are introduced and defined.

\subsection{Experimental setup}

This section describes the 3 different experiments performed in this paper, producing a total of 68 models. First, the level of corruption and the distribution of corruption is explored. Then, the impact of the length of text vs the number of observations is investigated. Finally, a comparison between models trained on synthetic data and real data is performed.

\subsubsection{Corruption level, and corruption distribution}

A fundamental question when using synthetic data to fine-tune an LM to learn how to do CLOCR-C is how much error is necessary and how should the error be distributed? This paper explores the topic using three experiments: first, by simply uniformly increasing the CER across the dataset; second, by varying the relationship between CER and WER; and finally, by combining CER and WER levels within the dataset. A more detailed description of the parameters are shown in the bullet points below. Given the extreme levels of corruption shown for CER = 0.5 shown in Table \ref{tab:corruption_results}, having a CER level much beyond 0.4 produces almost complete nonsense; as such, this paper will focus on the lower end of the CER scale.
The question this series of experiments seeks to answer is ``what corruption parameters provide the biggest improvement to model performance?"

\begin{enumerate}

    \item Data is corrupted with a fixed conditional probability such that the mean CER of the text is one of nine possible values, 0.1, 0.2, 0.3, 0.4, 0.5, 0.6, 0.7, 0.8, 0.9.
    \item Using a grid of WER-CER pairs, data is corrupted across a WER range of 0.1, 0.2, 0.3, 0.4, 0.5, 0.6, 0.7 for a target CER of 0.05, 0.1, 0.2, 0.3, 0.4. 
    \item Data is corrupted such that the dataset is constructed from a mixture of CER levels, where the samples are equally distributed between CER levels of  0-0.1, 0.1-0.2, 0.2-0.3, and 0.3-0.4. Each group makes up 25\% of the dataset. The WER level will be controlled by sampling from the empirical WER distribution for that CER group from the combination of the BLN600, CA and SMH datasets. A second model will be created the same as described but with uncorrupted data as a fifth group, each group making up 20\% of the dataset.
\end{enumerate}

This set of experiments will produce 46 models, 9 CER-only experiments, 35 CER-WER experiments, and 2 multi-CER-WER models.

It should be noted that as CER increases for a given WER, the relative corruption of the words being corrupted increases disproportionately. This means the document CER and the CER of the corrupted words may differ, which may play a role in the fine-tuned model's overall performance. One can say that the CER of the corrupted words is the ``effective CER" of the text. For the corruption function used in this paper, the effective CER saturates at 1.5. When CER is saturated, the solution to the problem is similar to masked language modelling, given the assumption that the text is all real words. However, a deeper exploration of these links is beyond the scope of this paper. More details on the relationship between CER-WER pairs and effective CER can be found in supplementary material in Figure 2.

\subsubsection{Text length vs number of observations}

Given that a CER-WER combination can be found that provides sufficient improvement in the test set, it is valuable to investigate whether other factors beyond the corruption type play a role in the performance of the LM at CLOCR-C. Two critical elements in the training of the LM are the length of the text supplied and the number of observations. This will be explored by modifying the total number of tokens in the training set and also the total number of tokens per observation. These two parameters will be varied according to Table \ref{tab:text_length_exp}. This experiment will produce 36 token text length pairs, from which we will gain insight into the importance of text length and token volume. An example of the different token lengths is shown in supplementary material section 3.

\begin{table}[h]
    \caption{The models will be trained using different combinations of text length and token budget. The table shows the number of data observations for a given observation token length and total token combination.}
    \centering
    \begin{tabular}{lrrrrr}
    \toprule
     & \multicolumn{5}{c}{Tokens per observation} \\
     Total Tokens& 200 & 100 & 50 & 25 & 10 \\
    \midrule
    1,638,400 & 8,192 & 16,384 & 32,768 & 65,536 & 163,840 \\
    819,200 & 4,096 & 8,192 & 16,384 & 32,768 & 81,920 \\
    409,600 & 2,048 & 4,096 & 8,192 & 16,384 & 40,960 \\
    204,800 & 1,024 & 2,048 & 4,096 & 8,192 & 20,480 \\
    102,400 & 512 & 1,024 & 2,048 & 4,096 & 10,240 \\
    51,200 & 256 & 512 & 1,024 & 2,048 & 5,120 \\
    25,600 & 128 & 256 & 512 & 1,024 & 2,560 \\
    \bottomrule
    \end{tabular}
    \label{tab:text_length_exp}
\end{table}

\subsubsection{Between dataset comparison}

Three models will be created, one for each of the datasets, BLN600, SMH, and CA. The performance of these models will be compared to synthetic data, the original dataset, the Base Llama3 model, and the current state-of-the-art Claude Opus. This comparison will provide insight into the relative performance of synthetic data vs available training data. The CA and SMH data have been transcribed with line breaks, making splitting the datasets easier; these two datasets will be split such that the median tokens per observation in the ground truth are 200, the same as the synthetic training set. As SMH and CA are small datasets, they will also be combined into a single Overproof dataset named after the project in which they were transcribed. The BLN600 data is not line aligned but has already been turned into a dataset using sequence matching \cite{thomas_leveraging_nodate}, however, the median tokens per sequence in the ground truth dataset is 26

\section{Results}
\label{sect:results}

\begin{figure}
    \centering
    \includegraphics[width=\linewidth]{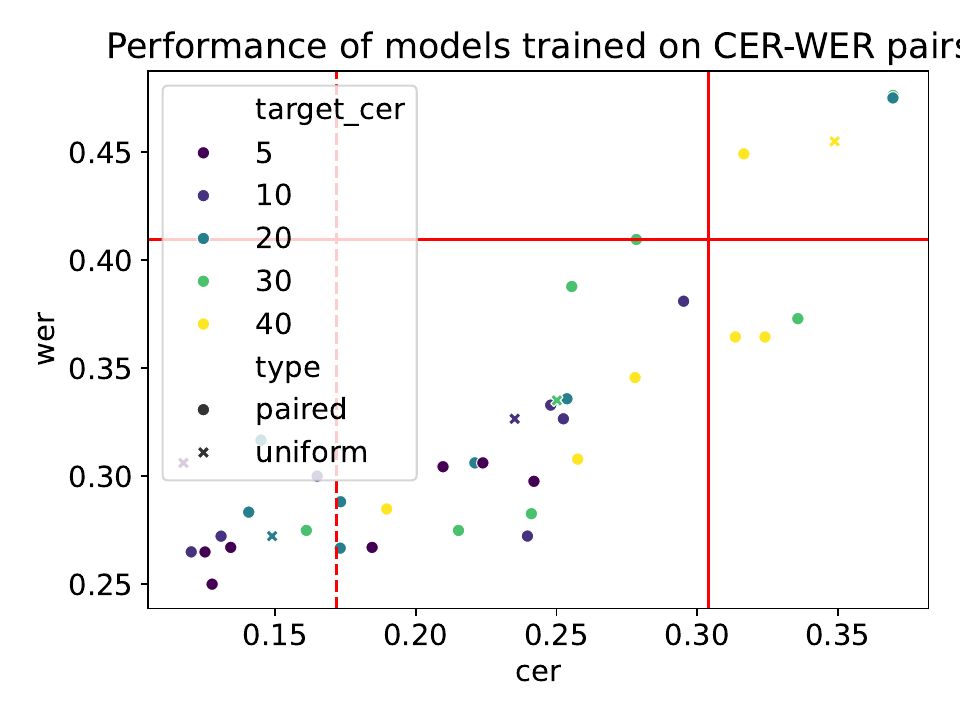}
    \caption{Solid red lines show the performance of the baseline Llama3 model, and the dashed red line shows the median CER of the NCSE dataset. Across the whole NCSE dataset, most models outperformed the baseline Llama3. Still, few managed to reduce the CER compared to the NCSE average of 0.17.}
    \label{fig:cer_wer_pairs}
\end{figure}

All models with uniform synthetic CER reduced the WER on the NCSE dataset. The CER was minimised when the synthetic training data had a CER of 0.2, close to the median CER of 0.17 of the NCSE dataset, as well as the 0.05 model. 
Figure \ref{fig:cer_wer_pairs} shows the performance of the models trained on synthetic data from different CER-WER combinations. In Figure \ref{fig:cer_wer_pairs}, the performance of the baseline Llama3 model is shown with a solid red lines whilst the original NCSE dataset average is shown with a dashed red line. The figure shows that whilst all models reduced error compared to the baseline WER and most models beat the Baseline Llama3, reducing the CER relative to the NCSE dataset was much more difficult. Primarily, models trained with low levels of CER were more likely to reduce the overall CER; however, there was not such a clear pattern with the WER values. A key factor of OCR errors is that the range of errors can be large and have considerably different impact on the text (see Table\ref{tab:corruption_results} for examples). The mean CER of the top ten models was 0.14 and the WER was 0.28; which is a an error reduction percentage of 21\% and 55\% respectively relative to the NCSE test dataset, and an improvement of CER \= 55\% and WER\=41\% relative to the Baseline Llama model.

\begin{figure}
    \centering
    \includegraphics[width=\linewidth]{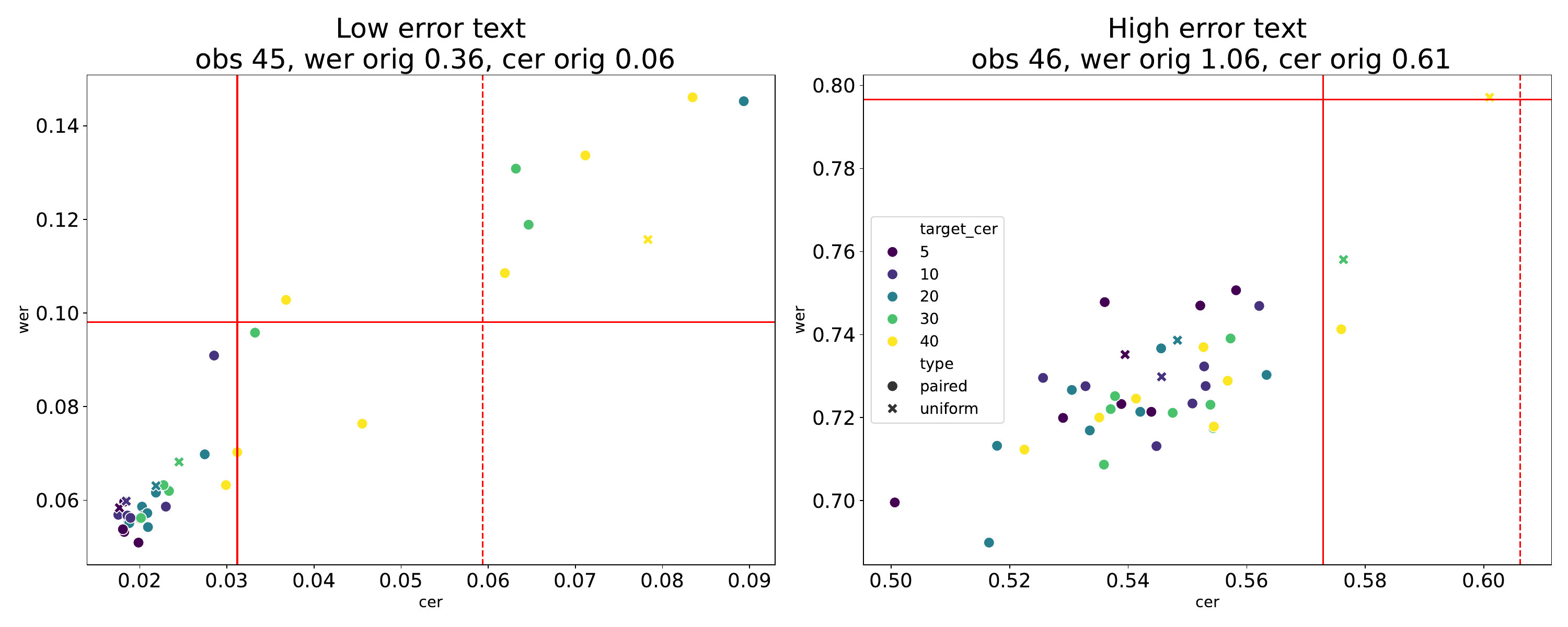}
    \caption{The figure shows the contrast in performance when looking at only the High (CER$>$0.17) and low (CER$\leq$0.17) corruption. The red lines show the performance of the base Llama model.}
    \label{fig:hi_lo_corr}
\end{figure}

The text was split at the median CER into high and low corruption subsets to explore the practical significance of the difference in overall CER on model performance. The High-low split was at CER = 0.17, shown in Figure \ref{fig:hi_lo_corr}. What Figure \ref{fig:hi_lo_corr} reveals is that there is a very large difference in the corruption, with the lower group having a median CER of 0.06 compared to the high corruption group's CER of 0.61. Models that perform well in the low corruption group all have very low target CER values, typically 0.05 or 0.1

In other words, the base Llama model increases the median error of the OCR data on average across the dataset except if the dataset is split into low and high corruption halves, in which case the base Llama model improves BOTH the high and low corruption sub-sets. This situation appears bizarre and reminiscent of Simpson's paradox \cite{simpson_interpretation_1951, blyth_simpsons_1972}. However, it can be explained by considering the CER distribution of the NCSE dataset as a Gaussian-mixture model of two components. When partitioned at the midpoint, the lower distribution is heavily right-skewed, with the centre of mass close to the lower limit, whilst the upper distribution is more normally distributed, with the centre of mass shifted away from the 0.17 lower limit. When viewed as a whole dataset, the mean CER of the base Llama3 model and the original dataset are both approximately 0.32, but their median values diverge.

In addition to the paradoxical appearance of model performance, the types of fine-tuned models that perform well are also very different. The low-corruption dataset was improved most by models trained on low-corruption data, whilst the high-corruption dataset was improved most by models trained on moderately corrupted data. This is interesting as it adds nuance to the findings of \cite{rychalska_models_2019}, who suggested that corruption should be at least as bad; these findings indicate that there may be an upper limit to corruption for successfully training models to correct OCR.

\FloatBarrier

\subsection{Tokens per observation and total tokens in dataset}

\begin{figure}
    \centering
    \includegraphics[width=\linewidth]{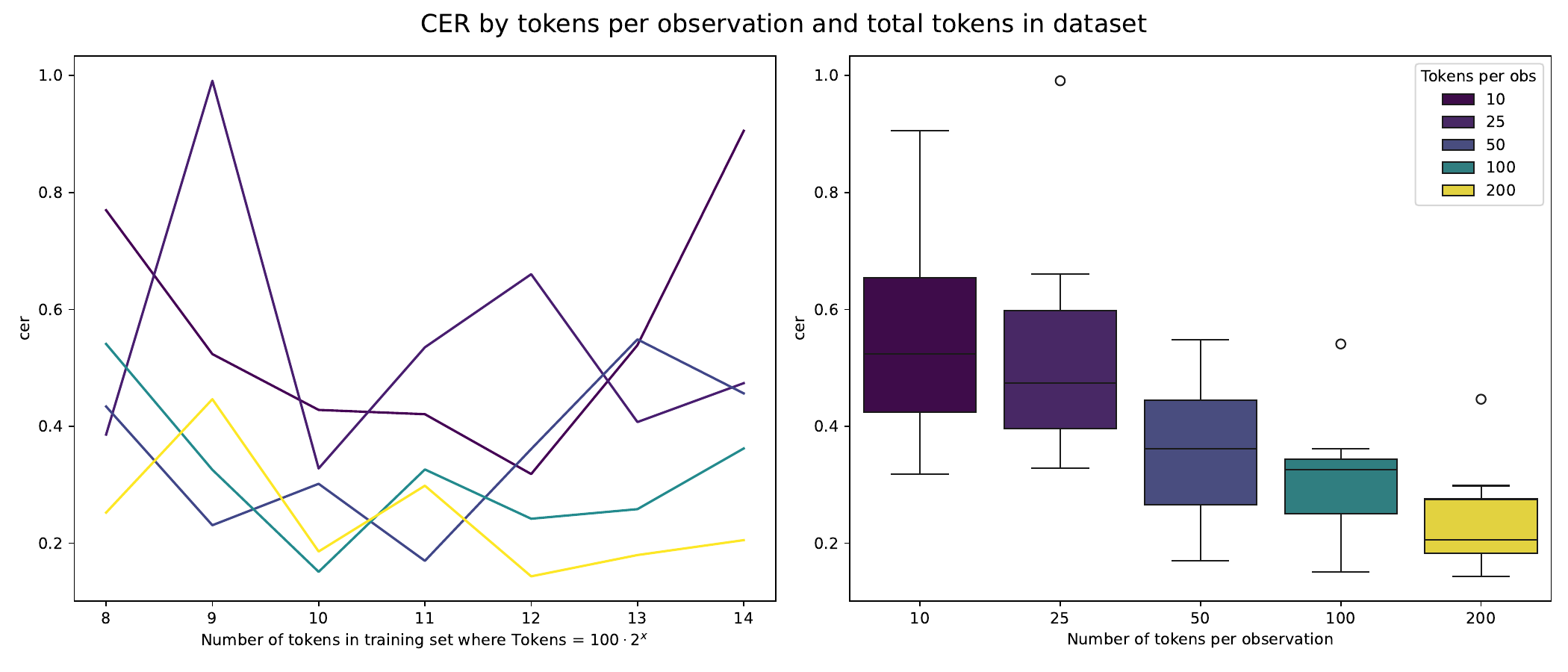}
    \caption{The figures shown here illustrate the interaction between the number of tokens per observation and the total number of tokens in the whole training set. Increasing the number of tokens in the dataset improves model performance. However, the impact of the total number of tokens can be difficult given the overall level of noise}
    \label{fig:length_volume}
\end{figure}

Given the results discussed in the previous section, the model trained with CER\=0.1 and WER = 0.2 was chosen as the base for the tokens-per-observations and tokens-in-training-set experiment.
The results of the experiment can be seen in Figure \ref{fig:length_volume} which shows the interaction between The two variables. Looking at the left panel of Figure \ref{fig:length_volume} the relationship between number of tokens and CER is difficult to see, it appears that there is substantial noise in the results, with only 200 tokens per observation models showing any discernible trend. Such an observation contrasts with the right panel, which shows how CER changes with increased tokens per observation. Here, the CER reduces up to the maximum of 200 tokens per observation.

\subsection{Comparing models trained on different datasets}

\begin{figure}
    \centering
    \includegraphics[width=\linewidth]{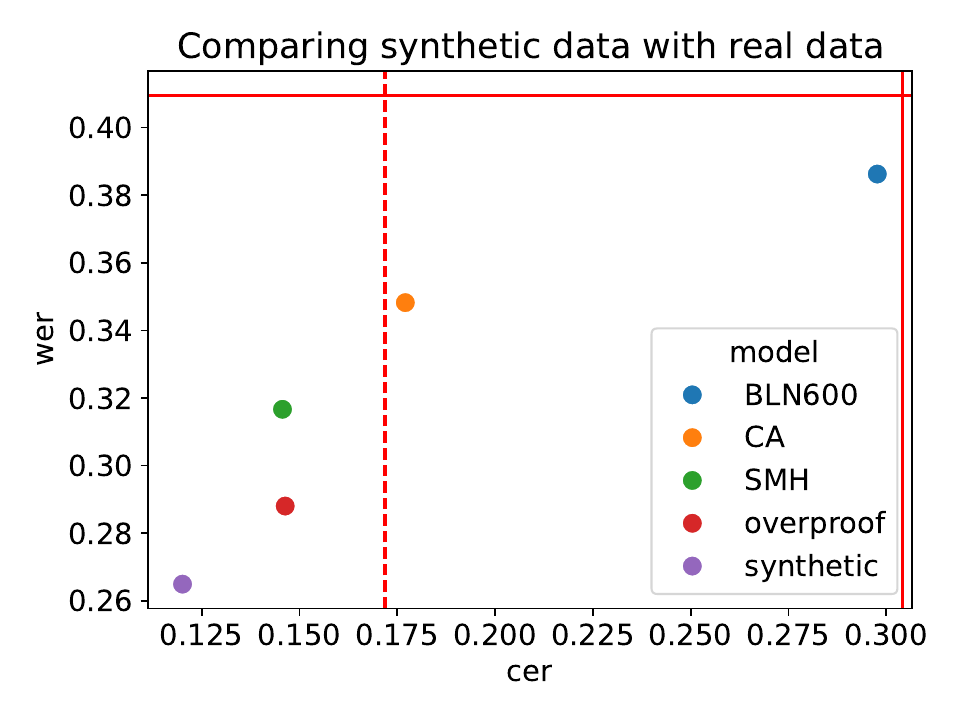}
    \caption{When comparing models the model trained on synthetic data against the models trained on several existing datasets, the advantages of the synthetic data are clear with a lower CER and WER.}
    \label{fig:compare_models}
\end{figure}

As shown in Figure \ref{fig:compare_models}, the synthetic data outperforms all models trained on real data, both in terms of WER and CER. In addition, it is worth noting that the BLN600 data did not perform much better than the baseline, which could be related to the short token lengths of the sequenced text. In addition, whilst the larger SMH dataset outperformed the CA dataset, the combined CA and SMH dataset was the best performing of the three but, interestingly, only really improved on WER, not CER. It should be noted that from \cite{bourne_clocr-c_2024}, the State of the art is Claude Opus, which had a CER of 0.07 and a WER of 0.15, substantially outperforming the synthetic data. Similar to the earlier analysis, the data was split into high and low corruption; the synthetic data performed best with the low corruption but dropped to third in the high corruption subset. In contrast, the BLN600 went from last on the low corruption subset to first on the high corruption subset, outperforming the base Llama by 8 percentage points. A figure showing the high-low split is shown in supplementary material 4.

\FloatBarrier
\section{Discussion}
\label{sect:disc}

The use of synthetic data for fine-tuning LM's to perform CLOCR-C has produced several findings. First, using synthetic data does improve the performance of an LM on real OCR data; the median performance of the top 10 models produced improved the CER of the base Llama model by 55\% and the WER by 32\%. Second, it is not only the amount of synthetic data that is important but also the distribution of the corruption in the data. Third, it is important to have enough tokens for each observation.

Beyond the main finding that synthetic data can be used to fine-tune LMs to perform CLOCR-C, it was interesting to observe how the data distribution played an important role in the final performance. Models with relatively low levels of corruption (CER$<0.2$) and WER in the low to mid-range ($0.2<$WER$<0.6$) seemed to perform well, with the best models reducing CER by 30\% and WER by 60\%. Part of this reason may be the different drivers of text correction for texts with low or high levels of corruption. For highly corrupted text, simply getting words right improves the WER. Still, substantial reductions may not be possible due to the lack of information in the text. In contrast, at low levels of corruption, the CER is improved by LM acting as what is effectively a contextual spell check; however, as the error rate is so low, a single incorrect word can have a large relative impact on overall performance. 

When training LMs, a general rule is that more diverse data leads to better-performing models. Therefore, it was surprising that the blended models had lackluster performance, with neither managing to beat the baseline median CER of 0.17. This unexpected result suggests that a better approach than training a single model on a mixture of data might be to train a mixture of experts model. Such an approach could work better as it would be able to explicitly route tasks to sub-models trained on data in the appropriate corruption range, sparking further curiosity and engagement in the field.

The exploration of the length of training context was revealing and supported the findings of \cite{dubey_llama_2024} and \cite{bourne_clocr-c_2024}, as models trained on observations with fewer tokens performed worse than those trained with more tokens, even though the total token training budget was fixed.

In this paper, the models were trained on text strings of 200 tokens; this is substantially less than most of the articles in the test set; it may be that, at inference time, some of the longer texts were corrected poorly due to the length itself and not the error distribution. However, such an investigation is beyond the scope of this paper. Another limitation of the paper is that the corruption function does not transpose entire words; this is a common feature of OCR errors. However, \verb|scrambledtext| could be combined with approaches such as WildNLP \cite{rychalska_models_2019} in a pipeline to include transposition.

Comparing the models showed how valuable synthetic data is, with the synthetic model outperforming all the real datasets. Whilst it can be said that synthetic data had an advantage in training on more tokens than the other models, this is one of the key advantages of synthetic data. The comparison also highlighted the challenge faced in creating real datasets, especially when using automatic sequencing, which is a common approach \cite{thomas_leveraging_nodate, chiron_icdar2017_2017, jiang_gutenberg-hathitrust_2021}. One interesting advantage that SMH and CA datasets was being line aligned; this allowed the data to be broken into almost arbitrary lengths of coherent text, something that is not possible with either \cite{bourne_clocr-c_2024} or \cite{booth_bln600_2024}. The fact that \verb|scrambledtext| can flexibly corrupt data independent of the amount of corruption in the text it was trained on shows the value of using a Markov model, which can be easily re-normalised to tune the corruption to the desired amount.

Although the synthetic data performed very well, it is still a way off beating current SOTA Claude Opus, which scored a CER of 0.07 and WER of 0.15, compared to the synth dataset's CER = 0.12 and WER 0.26 on the same dataset. A significant portion of the gap between Opus and the synthetically trained Llama model is the number of parameters, with Opus likely two orders of magnitude larger. Training and architecture could play a role, but as Opus is not an open-source model, exploring these differences is likely impossible.

One element of the process that could be improved to close the gap with the state of the art is the corruption function itself. The method described in Section \ref{sect:markov_model} is very simple and may explain the relatively poor performance compared to the other datasets in the high corruption subset. Perhaps a more sophisticated or nuanced approach may produce more realistic errors, improving model performance across the corruption range.

Although this paper shed light on how to train LMs to perform CLOCR-C it did not provide any insight into the mechanisms by which the LM's perform the task. The question is therefore still open as to whether the LM is essentially acting as a stochastic parrot \cite{bender_dangers_2021} or whether it is using a more fundamental understanding of language. This is an interesting question given the recent research by \textcite{kallini_mission_2024} into how LMs respond to impossible languages. It may be that one of the mechanisms used in CLOCR-C is that through fine-tuning, the LM gains some understanding of the structure of the ``impossible language" that is the OCR errors and learns to relate this to its already existing language representation and, in doing reveal the latent ``real language" masked by the errors. Further work could probe this aspect as advances could provide insight into LM language understanding and optimising for CLOCR-C.

Finally although this paper did not seek to optimise the performance the findings open the door for work exploring optimisation and model architecture comparison. This is made easier because, as is shown in Figure \ref{fig:length_volume}, the model trained on observations of 200 tokens still beats the baseline when trained on only 512 examples, roughly 5\% of the full dataset dramatically reducing training cost. Such work could then more concretely compare CLOCR-C with synthetic data against the state of the art.

\FloatBarrier
\section{Recommendations for training CLOCR-C models}
\label{sect:recs}
Given the results of the paper, below are some heuristics to consider when using synthetic data to train CLOCR-C models. 

\begin{enumerate}
    \item Know your data: Model performance is dependent on the corruption level of the data were trained on and are being applied to. Having an idea of the overall level of corruption in the data can help tune the synthetic data to maximise model performance.
    \item Corruption level: Train models on relatively low levels of corruption, a global CER of between 5-20\% should produce good results on most texts.
    \item Not enough corruption is better than too much: When creating synthetic data, training on overly corrupted data produces worse results than somewhat under-corrupted data.
    \item Corruption distribution: Concentrating corruption into fewer words generally gives better results than spreading the corruption more evenly; this may be because the effect is more similar to masked language modelling and so is more challenging. A CER-WER pair that gives an effective CER between 0.2 and 0.6 seems to perform well.
\item Synthetic text length: For a fixed token budget, having fewer observations containing more tokens provides better model performance than having more observations comprising fewer tokens.
    \item Amount of observations: Modern LMs are very well trained and don't need many examples; training on 1000 or fewer examples (low 100k tokens, not millions) is reasonable, possibly less than 500; however, with so few examples, the difference in training time will be a few minutes so better to err on the side of caution and, at least initially stick to a slightly higher number before working your way down if necessary.
    \item Drop heavily corrupted texts: Depending on the distribution of your data, it may be worth splitting the dataset into lightly and heavily corrupted datasets or dropping the heavily corrupted texts entirely. This is because models appear to specialise in one of the two types, and lightly corrupted models have a significantly larger relative increase in quality compared to heavily corrupted texts, which may be corrected to something semantically different to the original, causing issues in downstream analysis.
    \item BONUS! Line align your texts at transcription: Whilst not strictly part of synthetic data, it is still a valuable consideration. Line aligning means that text can be much more easily broken into sub-sequences without relying on automatic sequencing algorithms. Line-aligned text can also be used to split test-sets, creating more observations allowing better diagnostics on why models perform better or worse in various scenarios.
\end{enumerate}

\section{Conclusions}
\label{sect:concs}

This paper has shown that Language Models can be fine-tuned to perform Context Leveraging OCR Correction on entirely synthetic data. That is, training data from texts generated using a language model and then corrupted using a learned Markov corruption function. This has the potential to substantially reduce the time and cost necessary to perform an OCR correction project on a digital archive and supports the findings of previous work on training LMs with synthetic data \cite{gunasekar_textbooks_2023, li_textbooks_2023, tan_15-pints_2024} but applied to the OCR correction domain.
The experiments developed here resulted in 68 fine-tuned models and provided a comprehensive exploration of the parameter space, which gave sufficient insight into the behaviour of the LM trained using synthetic data to allow for the creation of 8 heuristics to guide practitioners when training a CLOCR-C model. The symbol agnostic, \verb|scrambledtext| package, makes it easy to generate synthetically corrupted training texts in any language. This paper indicates that the cost of fine-tuning a model of 8 Billion parameters is low, in the range of 10 cents to a few dollars, and the bulk of the cost of a recovery project will come from inference. As such, those seeking to recover corrupted archives should focus on efficient inference strategies. These strategies may involve taking advantage of advancements in model architecture \cite{shah_flashattention-3_2024, gu_mamba_2024, dao_transformers_2024, ma_era_2024} or hardware  \cite{sambanova_accelerated_2021,hall_training_2024, yang_optimized_2024, nvidia_nvidia_2024}.
Overall, it is hoped that the findings of this paper have made it easier to create CLOCR-C models and, as such, easier to recover quality archival texts in resource-constrained projects.
 perform

\section*{Data Availability}
The \verb|scrambledtext| library is available from \url{https://github.com/JonnoB/scrambledtext}, The synthetic articles and other data are available from the UCL data repository \cite{bourne_scrambled_2024}, the github repo of the main code is available from \url{https://github.com/JonnoB/scrambledtext_analysis}, a repo of code used for training the models is available from \url{https://github.com/JonnoB/training_lms_with_synthetic_data} and can be launched directly as a lightning.ai studio.

\printbibliography

\end{document}